\documentclass{article}

\usepackage{arxiv}

\usepackage[utf8]{inputenc} 
\usepackage[T1]{fontenc}    
\usepackage{hyperref}       
\usepackage{url}            
\usepackage{booktabs}       
\usepackage{amsfonts}       
\usepackage{nicefrac}       
\usepackage{microtype}      
\usepackage{lipsum}		
\usepackage{graphicx}
\usepackage{natbib}
\usepackage{doi}
\usepackage{amsmath}

\newcommand{\timevec}[1]{\mathbf{t2v}(#1)}
\newcommand{\emavec}[1]{\mathbf{e2v}(#1)}

\title{Learning Longitudinal Stress Dynamics from Irregular Self-Reports via Time Embeddings}

\date{} 					

\author{ {Louis Simon}\\
	Institut des Systèmes Intelligents et de Robotique\\
	  Sorbonne University\\
	Paris, France \\
	\texttt{louis.simon@isir.upmc.fr} \\
	\And
	{Mohamed Chetouani}\\
	Institut des Systèmes Intelligents et de Robotique\\
	  Sorbonne University\\
	Paris, France \\
	\texttt{mohamed.chetouani@isir.upmc.fr} \\
}

\renewcommand{\shorttitle}{Learning Longitudinal Stress Dynamics from Irregular Self-Reports via Time Embeddings}

\hypersetup{
pdftitle={A template for the arxiv style},
pdfsubject={q-bio.NC, q-bio.QM},
pdfauthor={David S.~Hippocampus, Elias D.~Striatum},
pdfkeywords={First keyword, Second keyword, More},
}

\begin{document}
\maketitle
\vspace{.5cm}

\begin{abstract}
The widespread adoption of mobile and wearable sensing technologies has enabled continuous and personalized monitoring of affect, mood disorders, and stress. When combined with ecological self-report questionnaires, these systems offer a powerful opportunity to explore longitudinal modeling of human behaviors. However, challenges arise from missing data and the irregular timing of self-reports, which make challenging the prediction of human states and behaviors. In this study, we investigate the use of time embeddings to capture time dependencies within sequences of Ecological Momentary Assessments (EMA). We introduce a novel time embedding method, Ema2Vec, designed to effectively handle irregularly spaced self-reports, and evaluate it on a new task of longitudinal stress prediction. Our method outperforms standard stress prediction baselines that rely on fixed-size daily windows, as well as models trained directly on longitudinal sequences without time-aware representations. These findings emphasize the importance of incorporating time embeddings when modeling irregularly sampled longitudinal data.
\end{abstract}

\let\thefootnote\relax\footnotetext{Accepted at ACII 2025 - Canberra}

\section{Introduction}
\begin{figure}[h]
    \centering
    \includegraphics[width=0.7\linewidth]{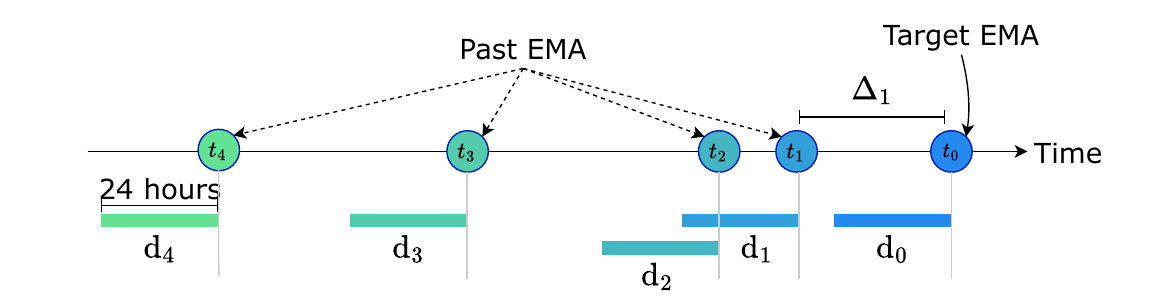}
    \caption{\textbf{Sequence of irregularly spaced self-report}: To predict target EMA at time $t_0$, we leverage daily sequences $\mathrm{d}_i$ aligned with past self-reports.}
    \label{fig:timeline}
\end{figure}

Considering the temporal dynamics of human affective states is essential for gaining meaningful insights into mood and related emotional conditions. In the case of automated stress detection, there is a need to develop computational models that can effectively capture these evolving patterns over time, in order to build systems that are both accurate and responsive. However, ecological assessments often result in irregularly spaced self-reports, as shown in Figure \ref{fig:timeline}. While the delays between reports may carry valuable information, processing such sequences remains challenging for many machine learning models. In the context of student stress prediction, researchers have explored the use of past labels, time to the next deadline, or time to the next label as forms of temporal information. These variables have typically been treated as simple covariates or incorporated into two-stage Bayesian prediction frameworks, as in \cite{mishra_continuous_2020}.

In our study, we propose to explicitly model temporal trends through the use of time embeddings. To capture delays between Ecological Momentary Assessments (EMAs) and gain insights into the dynamics of self-reports, we draw inspiration from positional encoding and time embedding techniques. Specifically, we leverage Time2Vec embeddings \cite{kazemi_time2vec_2019}, which are capable of encoding both periodic and non-periodic temporal patterns. We extend this representation to explicitly model longitudinal stress dynamics from irregular self-reports. This novel time embedding, which we call Ema2Vec, is designed to handle irregular EMA intervals and capture stress-related temporal dependencies.
\newpage
In this paper, we make the following contributions:
\begin{itemize}
    \item \textbf{Importance of longitudinal context:} We demonstrate that incorporating a longer historical context significantly improves model performance. 
    
    
  \item \textbf{Explicit representation of time through time embeddings:} We introduce time embeddings to explicitly represent temporal information, enabling more effective and lightweight modeling of temporal dynamics. 
  
    
    \item \textbf{Improvement of Ema2Vec over Time2Vec:} Ema2Vec, specifically designed to capture the dynamics of irregular self-reports, outperforms Time2Vec, confirming the advantage of using a task-specific temporal representation.
    
    \item \textbf{Forecasting without sensing data:} We show that the learned model is able to forecast future levels of self-reported stress based solely on historical self-reports, without the sensing data emitted during the day prior to the target EMA onset. 
\end{itemize}

This paper is organized as follows: Section \ref{sec:related} surveys past work on machine learning models for stress prediction and time embedding methods. Section \ref{sec:method} introduces the problem of longitudinal stress prediction along with the Ema2Vec time embedding module. Section \ref{sec:setup} describes the StudentLife dataset, the training process, as well as the stress forecasting evaluation task. Section \ref{sec:results} presents results on both longitudinal prediction and forecasting. Finally, Section \ref{sec:conclusion} concludes the paper by discussing the limitations of our approach and outlining directions for future work.

\section{Related Work}
\label{sec:related}

\subsection{Ecological Momentary Assessment}
The widespread use of smartphones and wearable devices has allowed research to continuously track behavioral signals in a environmental context. Researchers have extensively used smartphones and smartwatches to model stress, mood, and emotion based on phone log, position, and physiological signals. Data collection varies from few days \cite{kang_k-emophone_2023} to several month \cite{ben-zeev_crosscheck_2017}, target different population, e.g., students \cite{wang_studentlife_2014}, \cite{kang_k-emophone_2023} or hospital workers \cite{mundnich2020tiles}. These passive sensing apparatus are often paired with mobile applications specially developed for the gathering of Ecological Momentary Assessment (EMA) \cite{shiffman_ecological_2008} measures in the form of self-reported questionnaires, allowing for in-context probing of mental states. EMA surveys can be used to query users about their current affect \cite{miller_neuroticism_2009}, stress \cite{wang_studentlife_2014}, sleeping behaviors \cite{triantafillou_relationship_2019}, or mood disorders \cite{ebner-priemer_ecological_2009}. EMA-based studies are particularly suited for longitudinal data as they provide repeated measures under various context which can ultimately help to identify situations inducing changes.

In our study, we focus on the prediction of self-reported levels of stress. Researchers have been focusing on how different modalities, architectures, personalization techniques, and time windows can improve the prediction of self-reported stress levels using mobile sensing data. For example, Yu and Sano \cite{yu_semi-supervised_2023} proposed a semi-supervised model with active sampling to predict stress levels from physiological signal, leading to an improvement of the binary accuracy by 7.7\% to 13.8\% across three dataset. Bonafonte et al. \cite{bonafonte_analyzing_2023} conducted a thorough analysis on passive sensing modalities contribution and showed the importance WiFi features, which encode mobility patterns, and phone log features in binary and 3-class stress level classification. 

Beyond modalities, the choice of model architecture plays a crucial role. While many contributions rely on Multi-Layer Perceptrons (MLP), Long Short-Term Memory (LSTM) networks, or traditional machine learning models, researchers have also explored alternative architectures. For example, Harit et al. \cite{harit_monitoring_2024} proposed a Spatio-Temporal Graph Neural Network (STGNN) based model, demonstrating the potential of innovative architectures in this domain. Moreover, because of data heterogeneity and self-reports' obvious subjectivity, a significant effort towards model personalization is at the core of many studies. Jaques et al. \cite{jaques_predicting_2017} developed a multitask learning MLP and a Gaussian Process regression model with domain adaptation to enable personalized stress prediction in a population of 200 students. Shaw et al. \cite{shaw_personalized_2019} proposed a Multitask AutoEncoder model named CALM-Net and achieved an gain of 39.4 \% of the weighted-F1 compared to a non-personalized model. The aforementioned studies typically employ sensing sequences that capture context from the immediate past, spanning a day, or extending over longer periods of one to two weeks. Considering a longer temporal context can be beneficial, as stress induces changes at varying rates across different modalities. For instance, physiological signals may respond to stress within minutes or hours, while behavioral patterns captured through phone logs may evolve more gradually over several days. A major challenge in this context lies in distinguishing relevant from irrelevant information over extended periods of time. 

In our work, we address this by developing a more adaptive approach: we extract a discrete sequence of daily features preceding past self-reports and leverage the temporal information provided by the delays between reports using a dedicated time embedding module.

\subsection{Modeling temporal irregularity}

To effectively handle irregularly sampled time series, several architectures have been proposed to incorporate timestamp information into their internal representation.

Researchers have proposed a series of LSTM variants to better handle irregularly spaced sequences for tasks such as classification and forecasting. For instance, Time-LSTM \cite{zhu_what_2017} and Phased-LSTM \cite{neil_phased_2016} introduced modified gating mechanisms that incorporate temporal features directly into LSTM cells. These models have been applied to settings like user behavior modeling in recommender systems or processing event-based sensor data. Similarly, Baytas et al. \cite{baytas_patient_2017} proposed a time-aware LSTM for patient subtyping using Electronic Health Records (EHRs), where the elapsed time between medical events is used as an input to a modified LSTM cell. Later, Neural ODE \cite{chen_neural_2018} was proposed as an extension of RNNs to model continuous dynamics, natively enabling the modeling of irregular time series. Rubanova et al. \cite{rubanova_latent_2019} extended this work and proposed to combine RNN and Neural-ODE to better handle irregular time series with hidden states and continuous-time dynamics. While outperforming RNN-based approaches, Neural ODE models require a well specified dynamics system, which remain under-developed for psychological and behavioral stress process.

In contrast, attention-based model offer a greater flexibility by learning to attend to relevant events without the need of explicit dynamics modeling. Early implementation include RETAIN \cite{choi_retain_2016} which effectively incorporate RNN attention mechanism \cite{bahdanau_neural_2015} to learn an interpretable representation of irregularly sampled events from EHR. Further improvement was done through the ATTAIN architecture \cite{zhang_attain_2019} by explicitly including the RNN cell's context state as well as the timestamp information into the attention mechanism to better attend irregular sequences. This enhanced flexibility usually comes with higher computation complexity and costs, which is not suited for low data regime. 

In this work, we consider lightweight yet efficient methods, namely embeddings. Rather than modifying model architectures or introducing continuous-time dynamics, time embeddings encode temporal information directly into the input features. These representations —such as Time2Vec \cite{kazemi_time2vec_2019} or, to some extent, positional encodings \cite{vaswani_attention_2023, shaw_self-attention_2018}— allow neural network models to interpret relative or absolute timing without requiring explicit assumptions about system dynamics. For instance, Time2Vec learns a combination of periodic and linear components to embed timestamps, and has been adopted across a range of tasks, including mental health detection from irregular text data \cite{bucur_its_2023}.  While Time2Vec is effective in capturing periodic patterns within time series data, its applicability can be limited when the data does not exhibit such periodicity, motivating adaptations or extensions for broader use cases.

In this paper, we address this challenge by explicitly incorporating time gaps into a time-aware neural network architecture, enabling the model to better handle irregular intervals between self-reports and capture long-term dependencies in the feature space for stress prediction. Moreover, we propose a new set of activation functions tailored to EMA-based data.

\section{Method}
\label{sec:method}

\subsection{Problem statement}
\label{sec:problem_setting}

While RNN-based classifiers generally outperform fully connected models in human's states prediction tasks \cite{bonafonte_analyzing_2023}, scaling them to longer temporal contexts can be challenging—especially when working with small datasets. To address this, researchers often perform local aggregation of sensing data over time windows ranging from several minutes to hours. This typically involves extracting low-level features and applying statistical functionals — such as mean, standard deviation, minimum, and maximum — to these features \cite{Aigrain2018}. This approach not only helps reduce noise but also constrains the sequence length, enabling more efficient modeling. However, aggregation strategies based on statistical functionals are typically sensor-dependent, require careful feature engineering, and are sensitive to the chosen temporal horizon. In this work, to support modeling over extended timeframes —from daily to longitudinal scales — we extract a discretized sequence of day-level features aligned with the timestamps of previous self-reports of stress states. Unlike fixed time windows (e.g., minutes or hours), this discretization strategy aligns with the natural rhythm of daily life and self-report collection, enabling more meaningful temporal patterns to emerge while reducing sensitivity to irregular sampling and noise.

In this work, we demonstrate the relevance of our approach using the StudentLife dataset \cite{wang_studentlife_2014}, which contains data collected from 48 students at Dartmouth College over a period of 10 weeks. This dataset includes passive sensing data, such as sleep duration, conversation, physical activity, audio, GPS location, and phone event data like screen locking/unlocking, phone charging or light. Along these passive sensing signals, Ecological Momentary Assessment (EMA) were collected to probe students' well-being and mental health through questionnaires administered via a mobile app on a daily basis. Among the various questionnaires administered, students were asked to self-report their level of stress on a 1–5 Likert scale. During the study, they were prompted to complete an average of 8 EMAs (Ecological Momentary Assessments) per day and were free to respond at any time. In practice, students completed an average of 5.4 EMAs per day over the course of the study. Additionally, contextual covariates were collected, including sleep quality and duration, exam periods, and class attendance.

Formally, we consider a dataset of $N$ samples $\{(t^i, \mathbf{s}^i, c^i, y^i)\}_{i = 1, \dots, N}$, where $t^i$ denotes the time of an EMA report, $\mathbf{s}^i$ is the sequence of sensing signals starting one day prior to $t^i$, $c^i$ a vector representation of the sample's covariate, and $y^i$ is the corresponding self-reported stress level. We first aggregate each day-long sequence $\mathbf{s}^i$ into a daily representation $\mathrm{d}^i$. Given the limited size of our dataset, we define $\mathrm{d}^i$ using a set of statistical functionals over time: mean, maximum, minimum, standard deviation, median, sum, interquartile range, and mean crossing rate. While simple, this approach achieves good performance without the need for an additional encoder to process daily sequences. Previous contributions on the StudentLife dataset trained models to predict $y^i$ from continous sequences $\mathbf{s}^i$ ; we will refer to this problem as \textbf{standard stress prediction}. In this study, we introduce a new training and evaluation paradigm where input sequences are comprised of the past daily representations corresponding to the last $H$ self-reports; we denote this problem as \textbf{longitudinal stress prediction}. Namely, given a history length $H$, we construct a sequence of past daily representations $\mathbf{d}^i_{0:H} = [\mathrm{d}^i_0, \mathrm{d}^i_1, \dots, \mathrm{d}^i_H]$, where $\mathrm{d}^i_h$ corresponds to the day-level features associated with the $h$-th most recent EMA before $t^i$. We refer to this sequence as the \textbf{longitudinal sequence}, denoted $\mathbf{d}^i_{H:0}$. We also extract the corresponding sequence of absolute time delays $\Delta^i_{0:H}$ corresponding to the time difference between the target EMA onset with the preceding ones (see Figure~\ref{fig:timeline}).
\[
\Delta^i_{0:H} = [0, t^i_0 - t^i_1, \dots, t^i_0 - t^i_H] = [\Delta^i_0, \dots, \Delta^i_H].
\]

\begin{figure}[h]
    \centering
    \includegraphics[width=0.45\linewidth]{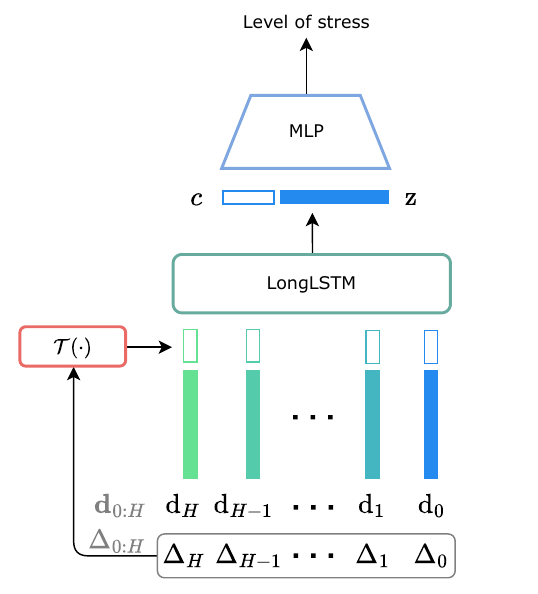}
    \caption{\textbf{LongLSTM + Time Embedding}: Our proposed model is composed of an LSTM processing a longitudinal sequence $\mathbf{d}_{0:H}$ concatenated with time delays $\Delta_{0:H}$ encoded by $\mathcal{T}$. The resulting hidden state $\mathrm{z}$ is combined with the covariate vector $c$ and fed to a Multi Layer Perceptron (MLP) for stress prediction.}
    \label{fig:archi}
\end{figure}

Given one ground truth self-reported level of stress $y^i$ from the EMA at time $t^i$, our goal is to learn a model $g(\cdot)$ from the longitudinal sequence $\mathbf{d}^i_{0:H}$ and the sequence of absolute time delay $\Delta^i_{0:H}$ (see Figure~\ref{fig:archi}). $\Delta^i_{0:H}$ is first encoded using a time embedding module $\mathcal{T}(\cdot)$; the resulting representation is then concatenated element-wise to $\mathbf{d}^i_{0:H}$. This contextualized sequence is then fed to an LSTM module which extracts a unique representation $\mathrm{z}^i$ for the pair $(\mathbf{d}^i_{0:H}, \Delta^i_{0:H})$:

\begin{equation}
    \mathrm{z}^i = \mathrm{LSTM}([\mathbf{d}^i_{0:H}, \mathcal{T}(\Delta^i_{0:H})])
\end{equation}

Finally, $\mathrm{z}^i$ is concatenated to covariate vector $c^i$ and fed to a Multi-Layer Perceptron with softmax activation for classification.
\begin{equation}
    \hat{y}^i = \mathrm{Softmax}(\mathrm{MLP}([\mathrm{z}^i, c^i])) = g(\mathbf{d}^i_{0:H}, \Delta^i_{0:H}, c^i)
\end{equation}
For simplicity, unless otherwise specified, we drop the sample index $i$ and refer to any sample sequence as $\mathbf{d}_{H:0}$.

\subsection{Self-report trends}
\label{sec:trends}
\begin{figure}[h]
    \centering
    \includegraphics[width=0.6\linewidth]{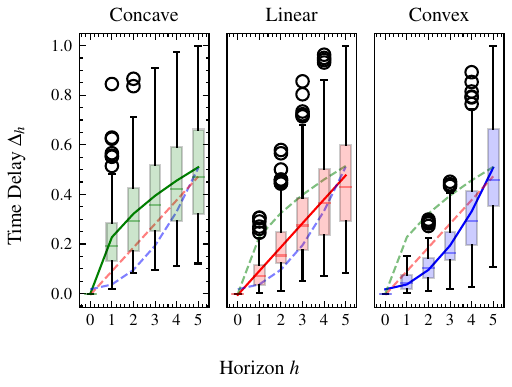}
    \caption{\textbf{Self-report trends}: For each class, average fitted functions (in plain lines) are plotted along other class average fit (in dotted lines) for comparison. Box-plots for each $\Delta_h$ are also reported.}
    \label{fig:trend_boxplot}
\end{figure}

Timing information at self-report onset, such as the time to a deadline, the day of the week, elapsed time since last report, can offer meaningful insights. This information is generally used as a covariate of the sequence of features. We argue that the delay between the past self-reports over a finite horizon $H$ also holds valuable information that can better contextualize sensing data within longitudinal sequences. 
We analyzed the function of absolute delays between a target EMA and the $H$ preceding ones, and identified three distinctive trends: linear, convex, and concave. We propose the following interpretation for classes of trends. 1) Linear trend indicates that the student has been consistent in their self-reporting, with past EMAs spaced at regular intervals. 2) Convex trend suggests that the student has not been consistent but is starting to report their stress more regularly. This pattern may reflect increasing academic pressure, such as exams or deadlines, or a stress-inducing period marked by heightened motivation or external demands. Finally, 3) Concave trend corresponds to samples where self-reports are becoming less frequent, potentially due to an approaching deadline, reduced engagement, or the onset of a significant change in stress levels. To determine which class each sample belongs to, we selected the best-fitting function among three candidates, defined as follows:
\begin{align}
    f_{linear}(\Delta_h) &= \alpha\times\Delta_h + \beta\\
    f_{convex}(\Delta_h) &= \alpha\times\Delta_h^2 + \beta\\
    f_{concave}(\Delta_h) &= \alpha\times\sqrt{\Delta_h} + \beta
\end{align}
For each sample in the dataset, we fit the best-matching function to the sequence $\Delta^i_{0:H}$ and classify it as exhibiting a linear, convex, or concave trend. The resulting distribution of trend types is reported in Figure \ref{fig:trend_boxplot}. While there is considerable variability in $\Delta_h$ as the distance from the target EMA increases, the three temporal trend classes remain clearly distinguishable. Such trends are difficult to distinguish directly within the daily functional sequences $\mathbf{d}_{0:H}$. Incorporating delays between irregularly spaced self-reports helps provide valuable temporal context to the sensing signals. To effectively integrate this information, we evaluate a well-established time embedding method, Time2Vec \cite{kazemi_time2vec_2019}, and introduce a novel approach called Ema2Vec.

\subsection{Time Embedding}
\label{sec:time2Vec}
\begin{figure}
    \centering
    \includegraphics[width=0.6\linewidth]{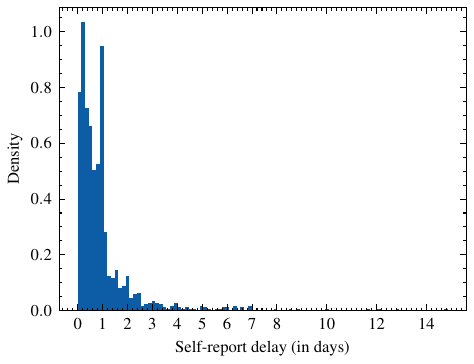}
    \caption{\textbf{Distribution of} $\Delta_h$: With an average of 5.4 self-reports a day across students, answered EMA are mostly spaced by least than a day. A small portion of self-reports were submitted  more than 2 days after the previous one.}
    \label{fig:delta_distrib}
\end{figure}

In order to both encode trends and positions of self-reports within longitudinal sequences, we encode absolute delays $\Delta = [\Delta_0, \dots, \Delta_H]$ with a time embedding module $\mathcal{T}(\cdot)$, concatenate the resulting representation to the corresponding daily function $\mathbf{d}_{0:H}$, and feed it to an LSTM. As shown in Figure \ref{fig:delta_distrib}, most self-reports are spaced by less than 3/4 days. Absolute delays above this threshold can be caused by application failure to register EMA, disengagement of student, or even stress inducing situations like exams. While this approach can provide valuable information for prediction, such delays occur infrequently and may be considered outliers. To reduce the effect of outliers and consider a fixed horizon, we set every $\Delta_h$ above $\Delta_{max} = $ 7 days to NaN prior to time encoding and use this to mask elements of $\mathbf{d}_{0:H}$ and $\Delta_{0:H}$. Each scalar $\Delta_h$ is then scaled to the range [0,1], with the maximum value $\Delta_h = 1$ corresponding to $\Delta_{\text{max}}$. 

\subsubsection{Time2Vec}

Time2Vec \cite{kazemi_time2vec_2019} is a simple yet efficient method for embedding any type of time information into vectors. For a given scalar notion of time $\tau$, $\timevec{\tau}$ is a $K+1$ embedding defined as follows:
\begin{equation}
    \timevec{\tau}[k] = \begin{cases}
        \omega_0\tau+ \varphi_0, \\
        \mathcal{F}(\omega_k\tau + \varphi_k), &\text{if} \; 1 \leq k \leq K
    \end{cases}
\end{equation}
where $\mathcal{F}$ is any periodic function, e.g., sine. Each dimension of $\timevec{\tau}$ is then scaled to [0,1] using a sigmoid function. In this study, we use the absolute delay from target EMA $\Delta_h$ as $\tau$.\\

\subsubsection{Ema2Vec}
\label{sec:ema2vec}
Building on the observations regarding self-report trends (figure \ref{fig:trend_boxplot}), we designed a time embedding module inspired by Time2Vec. Given an absolute time delay $\Delta_h$, the corresponding embedding $\mathbf{e}_{\Delta_h}$ is a $(K + 1)$-dimensional vector defined as:
\begin{equation}
    \emavec{\Delta_h}[k] = \begin{cases}
        a_0\Delta_h + b_0,  & \text{if} \; k = 0\\
        a_k(\Delta_h)^2 + b_k, &\text{if} \; 1 \leq k \leq K/2\\
        a_k\sqrt{\Delta_h} + b_k, &\text{if} \; K/2 < k \leq K\\
    \end{cases}
\end{equation}
The resulting embeddings are then normalized to unit norm, i.e., $||\mathbf{e}_{\Delta_h}||_2 = 1$.

\section{Experimental Setup}
\label{sec:setup}
\subsection{Dataset}

In this work, we evaluate our models using the StudentLife dataset \cite{wang_studentlife_2014}. There is no standardized splits of the StudentLife dataset for training machine learning models. Due to the small size of the dataset, the missing data, as well as the subjectivity of the self-reported labels, pre-processing methods and data cleaning procedures highly influence performance of models \cite{zhang_reproducible_2024}. For example, researchers often discard data coming from user with either too few or too much self-reported episodes in order to improve models' generalization capabilities. Full detailed descriptions of these procedures are often lacking in research papers and are only available in codes. Apart from the work of Luo et al.\cite{luo_dynamic_2024}, there is no available implementation of models and processing pipelines for the StudentLife dataset. Therefore, we use their preprocessed dataset\footnote{Available on \href{https://github.com/Information-Fusion-Lab-Umass/personalized-stress-prediction/releases/tag/processed_data_and_model_checkpoint}{Github}} and code for reproducing results. \\
In their study, Luo et al. build sequences by extracting hourly histograms of the activity, audio, conversation, phone charge, and phone sensing signals recorded in the past 24 hours before the EMA onset. As pointed out in the StudentLife dataset study \cite{wang_studentlife_2014}, deadlines are naturally stress inducing situations. For this reason, time to next deadline is concatenated to each element of the day-long sequences. Following \cite{mikelsons_towards_2017}, 1-5 levels of stress where transformed to a set of three classes, namely below median stress (0), median stress (1), and above median stress (2). Finally, the day of the week, the sleep rating and duration, as well as a boolean variable for indicating exam period are grouped to form a covariate vector for each samples. For additional details on the pre-processing steps, please refer to \cite{luo_dynamic_2024}. We add an additional step of pre-processing by computing the daily functional $\mathrm{d}_h$ from sequences of hourly histograms $\mathbf{s}$ and build longitudinal sequences accordingly. The final dataset is comprised of N = 1175 samples of size $(H+1)\times 76$.

\subsection{Tasks \& Baselines}
We evaluate our Ema2Vec module combined with an LSTM network to several baselines in three different settings:

\subsubsection{Standard Stress Prediction}
For the standard evaluation scheme, we compare our architecture with an LSTM trained with a day long sequence $\mathbf{s}$ comprised of 24 hourly histograms of the sensing signals. We report results both from our code and Luo et al. code with a slight modification. We noted that the original implementation of the baseline model used the test fold in a 5-fold cross-validation to select the best epoch, potentially leading to overestimated performances. To avoid such data contamination, we modified the code to include a separate validation split for epoch selection.

\subsubsection{Longitudinal Stress Prediction}
We then compare our approach with models taking longitudinal sequences $\mathbf{d}_{0:H}$ as inputs, as described in section \ref{sec:problem_setting}. We report the performances of a vanilla LSTM, denoted as LongLSTM, followed by its variant with a Time2Vec for time encoding. To demonstrate the ability of Ema2Vec and Time2Vec to accurately encode temporal dependencies on their own, we report the performance of the LongLSTM model with the sequence of delays ${\Delta}_{0:H}$ concatenated to the input features $\mathbf{d}_{0:H}$. We refer to this variant as TimeConcat LongLSTM. Results of both standard and longitudinal stress are reported in \ref{sec:long_pred}.

\subsubsection{Stress Forecasting}
\label{sec:task}
Models trained on longitudinal stress prediction utilize a discrete sequence of daily feature aligned to past self-report, which requires students to answer EMA; this ultimately reduces the relevance of our model in an interventional scenario where we want to be able to alert students about future stress events. To demonstrate the efficiency of our model in such scenario, we use models trained on longitudinal stress prediction and evaluate them on a \textbf{stress forecasting} task, without further finetuning. Formally, we predict stress level $y^i$ at time $t^i$ using only daily features $\mathbf{d}^i_{1:H}$ aligned with past self-reports, i.e., by excluding target EMA daily feature $\mathrm{d}^i_0$. To match the dimension of the time input feature $\Delta_{0:H+1}$, the longitudinal sequence is padded with the daily feature $\mathrm{d}^i_1$ of the immediate past report. The stress forecasting is therefore described as :

\begin{equation}
     \hat{y}^i = g(\mathbf{\tilde{d}}^i_{0:H+1}, \Delta^i_{0:H+1}, c^i)
\end{equation}

with $\mathbf{\tilde{d}}^i_{0:H+1} = [\mathrm{d}^i_1 , \mathrm{d}^i_1,  \mathrm{d}^i_2 , \dots, \mathrm{d}^i_H ]$. We report results of the stress forecasting task in section \ref{sec:forecast}.

\subsection{Training and Evaluation}
In order to avoid data leakage from the test set to the train set and guarantee consecutive daily self-report in our longitudinal model, we conduct evaluation with a 5-fold chronological cross validation. Each self-reported EMA is sorted chronologically per student and divided in 5 splits as a way to equally evaluate across each and every individual. The resulting splits corresponds to five temporally distinct periods equivalent to 2 weeks. We perform grid search for hyperparameter tuning by maximizing the weighted F1-score on a portion of the training folds data, ending up with a 70\%/10\%/20\% train/validation/test split. Due to the label imbalance, it is easy to reach a decent weighted F1-score by always predicting one of the two majority classes while performing around random choice in macro F1-score. We choose to report both the macro and weighted F1-score to have an unbiased estimation of the performances. Regarding the training configuration, we set the number of past self-reports H to 4, corresponding to an average horizon of $4 \pm 2$ days. The LSTM model was configured with a hidden size of 128, followed by dense layers with sizes 64, 32, and 3. To mitigate overfitting, a weight decay of $ 5 \times 10^{-5} $ was applied, along with dropout rates of 0.3, 0.1, and 0.2 for the LSTM and the first two dense layers, respectively. This configuration was maintained for both the Time2Vec and Ema2Vec variants, with the number of activations in the time embedding module $ \mathcal{T} $ set to 8 and 9, respectively. The learning rate for the LongLSTM was set to $ 2 \times 10^{-5} $, while a learning rate of $ 5 \times 10^{-4} $ was used for $ \mathcal{T} $, both with the Adam optimizer. A batch size of 4 was employed throughout training.



\section{Results}
\label{sec:results}

\subsection{Standard \& Longitudinal stress prediction}
\label{sec:long_pred}
Results of the 5-fold chronological cross-validation for both the standard and longitudinal stress prediction tasks are presented in Table \ref{fig:res_long_stress}. Overall, the LongLSTM model and its variants consistently outperform the LSTM baseline trained on standard stress prediction, with relative performance improvements ranging from 5.6\% to 8.5\% in weighted F1-score. These findings suggest that modeling longer sequences, even when irregularly spaced, ultimately benefits performances. Among the longitudinal models, incorporating a time embedding module—either Time2Vec or Ema2Vec—leads to further performance gains of 1.9\% and 2.9\%, respectively. This indicates that explicitly encoding temporal information contributes to model effectiveness. Furthermore, the performance of the TimeConcat LongLSTM variant remains closer to that of the vanilla LongLSTM, suggesting that the learned time embeddings in Ema2Vec and Time2Vec offer a more expressive and nuanced representation of temporal dynamics compared to a model that simply concatenates features over time. This highlights the advantages of explicitly modeling time through embeddings, allowing the model to better capture complex temporal patterns and dependencies in the data. Finally, results indicate that the Ema2Vec module slightly outperforms Time2Vec, with a relative improvement of 1\%. This suggests that designing the activation functions of the time embedding mechanism to better capture self-report temporal trends leads to richer, more data-specific temporal representations. By tailoring the time embeddings to the characteristics of the data, Ema2Vec is able to capture subtle temporal dynamics more effectively than Time2Vec, improving performance in longitudinal stress prediction.

\begin{table*}[h]
\centering
\begin{tabular}{|l|l|c|c|}
\hline
Model      & Input                    & F1 Macro          & F1 Weighted      \\ \hline
Oracle & None  & $0.504 \pm 0.218$ & $0.646 \pm 0.196$\\ \hline
LSTM (reproduced from \cite{luo_dynamic_2024})     & $\mathbf{s}$  & $0.316 \pm 0.040$ & $0.378 \pm 0.044$ \\ \hline
LSTM       & $\mathbf{s}$             & $0.369 \pm 0.032$ & $0.406 \pm 0.045$\\ \hline
LongLSTM & $\mathbf{d}_{0:H}$    & $0.438 \pm 0.050$ & $0.462 \pm 0.057$\\ \hline
TimeConcat LongLSTM & $\mathbf{d}_{0:H}$   & $0.442 \pm 0.054$  & $0.465 \pm 0.055$\\ \hline
LongLSTM + Time2Vec &  $\mathbf{d}_{0:H}$   & $0.461 \pm 0.044$ & $0.481 \pm 0.044$\\ \hline
LongLSTM + Ema2Vec & $\mathbf{d}_{0:H}$     & $\mathbf{0.471 \pm 0.042}$ & $\mathbf{0.491 \pm 0.043}$\\ \hline
\end{tabular}
\caption{\textbf{5-fold Chronological Cross Validation classification performance} (Average $\pm$ Standard Deviation)} 
\vspace{.2cm}

\label{fig:res_long_stress}
\end{table*}

\subsection{Stress forecasting}
\label{sec:forecast}
We evaluate models trained for longitudinal stress prediction on a subsequent stress forecasting task, as described in Section \ref{sec:task}, and report the results in Table \ref{fig:forecast_results}. It is important to note that the models were not fine-tuned for this task. Our proposed time embedding module, Ema2Vec, outperforms other longitudinal models, namely LongLSTM with and without Time2Vec. Furthermore, the drop in weighted F1-score from the regular longitudinal stress prediction task is smaller for Ema2Vec, with a $-1.8\%$ reduction compared to $-2.3\%$ and $-3.2\%$ for LongLSTM + Time2Vec and LongLSTM, respectively. Surprisingly, the standard deviation across the 5 folds for the LongLSTM + Ema2Vec model is reduced to $3.2\%$, indicating that the model exhibits strong generalization capabilities when forecasting future stress levels. These results suggest that our model is capable of effectively leveraging the learned time embedding representation to forecast future stress levels, even in the absence of immediate past sensing data.

\begin{table}[h]
\centering
\begin{tabular}{|l|l|l|}
\hline
Model      & F1 Macro          & F1 Weighted      \\ \hline
LongLSTM &  $0.411 \pm 0.052$ & $0.430 \pm 0.056$ \\ \hline
LongLSTM + Time2Vec &  $0.437 \pm 0.044 $& $0.458 \pm 0.052$ \\ \hline
LongLSTM + Ema2Vec &  $\mathbf{0.454 \pm 0.029}$ & $\mathbf{0.473 \pm 0.032}$\\ \hline
\end{tabular}
\vspace{.2cm}
\caption{\textbf{Stress Forecasting Results}}
\label{fig:forecast_results}
\end{table}

\subsection{Analysis of Ema2Vec embeddings}
Beyond predictive performance, the Ema2Vec module exhibits interpretable behavior with respect to temporal self-report patterns. For each trend class, we analyze the similarity profile defined as $\mathrm{sim}(0, h) = \cos(\emavec{\Delta_0}, \emavec{\Delta_h})$, for $h = 0, \dots, H$, which measures the cosine similarity between the time embedding of the target self-report $\emavec{\Delta_0}$ and that of a past report at position $h$. A similarity $\mathrm{sim}(0, h)$ close to 1 indicates that the embedding at lag $h$ is nearly identical to the target time embedding.

\begin{figure}[h]
    \centering
    \includegraphics[width=0.55\linewidth]{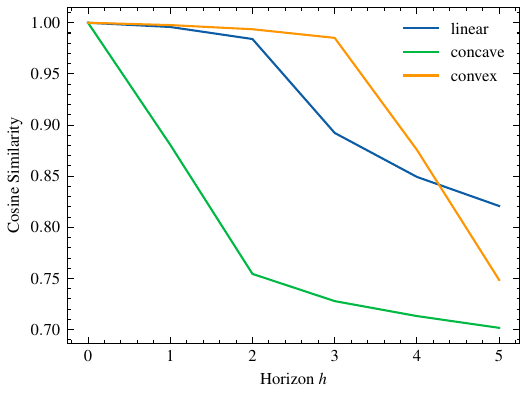}
    \caption{\textbf{Ema2Vec similarity profiles}: We report cosine similarity between target EMA Ema2Vec embedding $\emavec{\Delta_0}$ and the previous $h$ embeddings $\emavec{\Delta_h}$. Average similarity profile is reported for each class of trends.}
    \label{fig:time_sim}
\end{figure}

After assigning each sample to one of three temporal trend classes—linear, convex, or concave—we compute and report the average similarity profile for each group in Figure \ref{fig:time_sim}.

For linear trends, the embeddings corresponding to the most recent reports, $\emavec{\Delta_1}$ and $\emavec{\Delta_2}$, are highly similar to the target embedding $\emavec{\Delta_0}$, followed by a gradual decline, reaching approximately $\mathrm{sim}(0, H) = 0.82$. In the case of convex trends, the embeddings remain highly similar up to $h = 3$, after which there is a sharp decline in similarity, reaching $\mathrm{sim}(0, H) = 0.75$. This pattern is consistent with sequences where participants begin reporting regularly after a long period of inactivity; the earlier self-reports (e.g., at $h = H$ or $H{-}1$) differ significantly from the target report. Lastly, for concave trends, the similarity profile exhibits a sharp linear decrease up to \( h = 2 \), followed by a more gradual decline, with similarity reaching \( \mathrm{sim}(0, H) = 0.70 \). This pattern suggests that the time embedding for the target report \( \emavec{\Delta_0} \) becomes increasingly dissimilar to earlier reports, particularly those beyond the second most recent, indicating that Ema2Vec captures a temporally localized encoding. Overall, the observed similarity curves demonstrate that Ema2Vec effectively captures meaningful temporal dynamics and adapts to the specific patterns of self-report trends, supporting its ability to improve longitudinal stress prediction.

\section{Conclusion}
\label{sec:conclusion}
In this study, we introduced a novel training and evaluation scheme for stress prediction, termed longitudinal stress prediction. Unlike traditional approaches that use a fixed temporal window (e.g., one day), we proposed to leverage longitudinal sequences composed of daily representations of sensing data, aligned with past self-reports. To model the temporal dependencies in these irregularly spaced Ecological Momentary Assessments (EMAs), we proposed a new time embedding module, Ema2Vec. When combined with an LSTM and trained on longitudinal sequences, our model systematically outperforms other baselines  in predicting stress over time. Notably, we demonstrated that our Ema2Vec-based model, trained on longitudinal stress prediction, can infer stress labels ahead without additional finetuning.

There is a growing interest in the Affective Computing community for longitudinal datasets, particularly in applications such as interventional therapy or continuous monitoring of affective state with wearables. We believe that learning data-specific time embeddings, as done with Ema2Vec, has broader applicability in other contexts that involve EMA or repeated, irregularly spaced measurements. However, it is important to acknowledge that such approach heavily relies on data. In our study, we designed our Ema2Vec with dataset-specific insights into self-reporting trends. As such, generalization to other datasets or settings may be limited in the absence of clear temporal patterns. In future works, we aim to develop more flexible time embeddings that require minimal prior knowledge of the data. To this end, embeddings based on learnable functions such as splines could adapt better to diverse datasets. With access to more data, we also plan to combine Ema2Vec with more expressive architectures like Transformers. We hope this work encourages further research into learning time embedding modules to better model irregularly spaced longitudinal sequences.

\section*{Ethical Impact Statement}
This study uses the StudentLife dataset, a publicly available and anonymized collection of behavioral and self-report data gathered from 48 Dartmouth College students. Data were collected with the informed consent of participants, and identifying information was removed prior to public release. Although the data set provides rich longitudinal insights into behavioral patterns and includes a sufficient number of participants to evaluate our model, it is important to acknowledge that all individuals in the dataset belong to a relatively homogeneous group, namely college students from a single institution in the United States. As such, our findings may not be generalized to other populations with different demographic, cultural, or lifestyle factors. In this work, we propose a new training paradigm for modeling stress from irregularly sampled sequences of passive-sensing data. Unlike standard stress prediction approaches that rely on fixed daily windows, our method captures behavioral trends over extended periods (up to a week), offering a more comprehensive view of longitudinal stress dynamics. However, the use of long-term behavioral data comes with additional privacy concerns, particularly when dealing with sensitive mental health information. Finally, we emphasize that our model is not intended for diagnostic use or as a replacement for mental health professionals. We advocate for a cautious interpretation of our results, especially in clinical or applied settings, and recommend that future work validates such models in more diverse populations and in collaboration with mental health experts to ensure fairness and ethical implementation. 

\section*{Acknowledgement}
\noindent This work was supported by ERA PerMed JTC 2019 “Personalised Medicine: Multidisciplinary Research Towards Implementation” under the ERANET Cofund Action on PM - N° 779282 and ANR.

\bibliographystyle{plainnat}
\bibliography{biblio}  

\begin{thebibliography}{30}
\providecommand{\natexlab}[1]{#1}
\providecommand{\url}[1]{\texttt{#1}}
\expandafter\ifx\csname urlstyle\endcsname\relax
  \providecommand{\doi}[1]{doi: #1}\else
  \providecommand{\doi}{doi: \begingroup \urlstyle{rm}\Url}\fi

\bibitem[Aigrain et~al.(2018)Aigrain, Spodenkiewicz, Dubuisson, Detyniecki, Cohen, and Chetouani]{Aigrain2018}
Jonathan Aigrain, Michel Spodenkiewicz, Séverine Dubuisson, Marcin Detyniecki, David Cohen, and Mohamed Chetouani.
\newblock Multimodal stress detection from multiple assessments.
\newblock \emph{IEEE Transactions on Affective Computing}, 9\penalty0 (4):\penalty0 491--506, 2018.
\newblock \doi{10.1109/TAFFC.2016.2631594}.

\bibitem[Bahdanau et~al.(2015)Bahdanau, Cho, and Bengio]{bahdanau_neural_2015}
Dzmitry Bahdanau, Kyunghyun Cho, and Yoshua Bengio.
\newblock Neural {Machine} {Translation} by {Jointly} {Learning} to {Align} and {Translate}.
\newblock In Yoshua Bengio and Yann LeCun, editors, \emph{3rd {International} {Conference} on {Learning} {Representations}, {ICLR} 2015}, 2015.

\bibitem[Baytas et~al.(2017)Baytas, Xiao, Zhang, Wang, Jain, and Zhou]{baytas_patient_2017}
Inci~M. Baytas, Cao Xiao, Xi~Zhang, Fei Wang, Anil~K. Jain, and Jiayu Zhou.
\newblock Patient {Subtyping} via {Time}-{Aware} {LSTM} {Networks}.
\newblock In \emph{Proceedings of the 23rd {ACM} {SIGKDD} {International} {Conference} on {Knowledge} {Discovery} and {Data} {Mining}}, pages 65--74, Halifax NS Canada, August 2017. ACM.
\newblock ISBN 978-1-4503-4887-4.
\newblock \doi{10.1145/3097983.3097997}.

\bibitem[Ben-Zeev et~al.(2017)Ben-Zeev, Brian, Wang, Wang, Campbell, Aung, Merrill, Tseng, Choudhury, Hauser, Kane, and Scherer]{ben-zeev_crosscheck_2017}
Dror Ben-Zeev, Rachel Brian, Rui Wang, Weichen Wang, Andrew~T. Campbell, Min S.~H. Aung, Michael Merrill, Vincent W.~S. Tseng, Tanzeem Choudhury, Marta Hauser, John~M. Kane, and Emily~A. Scherer.
\newblock {CrossCheck}: {Integrating} self-report, behavioral sensing, and smartphone use to identify digital indicators of psychotic relapse.
\newblock \emph{Psychiatric rehabilitation journal}, 40\penalty0 (3):\penalty0 266--275, September 2017.
\newblock ISSN 1095-158X.
\newblock \doi{10.1037/prj0000243}.

\bibitem[Bonafonte et~al.(2025)Bonafonte, Bustos, Larrazolo, Luna, Arenas, Baro, Tourgeman, Balcells, and Lapedriza]{bonafonte_analyzing_2023}
Irene Bonafonte, Cristina Bustos, Abraham Larrazolo, Gilberto Lorenzo~Martinez Luna, Adolfo~Guzman Arenas, Xavier Baro, Isaac Tourgeman, Mercedes Balcells, and Agata Lapedriza.
\newblock Analyzing the contribution of different passively collected data to predict {Stress} and {Depression}.
\newblock \emph{2023 11th International Conference on Affective Computing and Intelligent Interaction Workshops and Demos (ACIIW)}, March 2025.
\newblock \doi{10.1109/ACIIW59127.2023.10388089}.

\bibitem[Bucur et~al.(2023)Bucur, Cosma, Rosso, and Dinu]{bucur_its_2023}
Ana-Maria Bucur, Adrian Cosma, Paolo Rosso, and Liviu~P. Dinu.
\newblock It’s {Just} a {Matter} of {Time}: {Detecting} {Depression} with {Time}-{Enriched} {Multimodal} {Transformers}.
\newblock In \emph{Advances in {Information} {Retrieval}: 45th {European} {Conference} on {Information} {Retrieval}, {ECIR}}, pages 200--215, 2023.
\newblock ISBN 978-3-031-28243-0.
\newblock \doi{10.1007/978-3-031-28244-7_13}.

\bibitem[Chen et~al.(2018)Chen, Rubanova, Bettencourt, and Duvenaud]{chen_neural_2018}
Ricky T.~Q. Chen, Yulia Rubanova, Jesse Bettencourt, and David~K Duvenaud.
\newblock Neural {Ordinary} {Differential} {Equations}.
\newblock In \emph{Advances in {Neural} {Information} {Processing} {Systems}}, volume~31. Curran Associates, Inc., 2018.

\bibitem[Choi et~al.(2016)Choi, Bahadori, Sun, Kulas, Schuetz, and Stewart]{choi_retain_2016}
Edward Choi, Mohammad~Taha Bahadori, Jimeng Sun, Joshua Kulas, Andy Schuetz, and Walter Stewart.
\newblock {RETAIN}: {An} {Interpretable} {Predictive} {Model} for {Healthcare} using {Reverse} {Time} {Attention} {Mechanism}.
\newblock In \emph{Advances in {Neural} {Information} {Processing} {Systems}}, volume~29. Curran Associates, Inc., 2016.

\bibitem[Ebner-Priemer and Trull(2009)]{ebner-priemer_ecological_2009}
Ulrich~W. Ebner-Priemer and Timothy~J. Trull.
\newblock Ecological momentary assessment of mood disorders and mood dysregulation.
\newblock \emph{Psychological Assessment}, 21\penalty0 (4):\penalty0 463--475, 2009.
\newblock ISSN 1939-134X.
\newblock \doi{10.1037/a0017075}.

\bibitem[Harit et~al.(2024)Harit, Sun, Yu, and Moubayed]{harit_monitoring_2024}
Anoushka Harit, Zhongtian Sun, Jongmin Yu, and Noura~Al Moubayed.
\newblock Monitoring {Behavioral} {Changes} {Using} {Spatiotemporal} {Graphs}: {A} {Case} {Study} on the {StudentLife} {Dataset}.
\newblock In \emph{NeurIPS 2024 Workshop on Behavioral ML}, October 2024.

\bibitem[Jaques et~al.(2017)Jaques, Rudovic, Taylor, Sano, and Picard]{jaques_predicting_2017}
Natasha Jaques, Ognjen Rudovic, Sara Taylor, Akane Sano, and Rosalind Picard.
\newblock Predicting {Tomorrow}’s {Mood}, {Health}, and {Stress} {Level} using {Personalized} {Multitask} {Learning} and {Domain} {Adaptation}.
\newblock In \emph{Proceedings of {IJCAI} 2017 {Workshop} on {Artificial} {Intelligence} in {Affective} {Computing}}, pages 17--33. PMLR, September 2017.
\newblock ISSN: 2640-3498.

\bibitem[Kang et~al.(2023)Kang, Choi, Park, Cha, Kim, Khandoker, Hadjileontiadis, Kim, Jeong, and Lee]{kang_k-emophone_2023}
Soowon Kang, Woohyeok Choi, Cheul~Young Park, Narae Cha, Auk Kim, Ahsan~Habib Khandoker, Leontios Hadjileontiadis, Heepyung Kim, Yong Jeong, and Uichin Lee.
\newblock K-{EmoPhone}: {A} {Mobile} and {Wearable} {Dataset} with {In}-{Situ} {Emotion}, {Stress}, and {Attention} {Labels}.
\newblock \emph{Scientific Data}, 10\penalty0 (1):\penalty0 351, June 2023.
\newblock ISSN 2052-4463.
\newblock \doi{10.1038/s41597-023-02248-2}.
\newblock Publisher: Nature Publishing Group.

\bibitem[Kazemi et~al.(2019)Kazemi, Goel, Eghbali, Ramanan, Sahota, Thakur, Wu, Smyth, Poupart, and Brubaker]{kazemi_time2vec_2019}
Seyed~Mehran Kazemi, Rishab Goel, Sepehr Eghbali, Janahan Ramanan, Jaspreet Sahota, Sanjay Thakur, Stella Wu, Cathal Smyth, Pascal Poupart, and Marcus Brubaker.
\newblock {Time2Vec}: {Learning} a {Vector} {Representation} of {Time}.
\newblock July 2019.
\newblock \doi{10.48550/arXiv.1907.05321}.
\newblock arXiv:1907.05321 [cs].

\bibitem[Luo et~al.(2024)Luo, Deznabi, Shaw, Simsiri, Rahman, and Fiterau]{luo_dynamic_2024}
Yunfei Luo, Iman Deznabi, Abhinav Shaw, Natcha Simsiri, Tauhidur Rahman, and Madalina Fiterau.
\newblock Dynamic clustering via branched deep learning enhances personalization of stress prediction from mobile sensor data.
\newblock \emph{Scientific Reports}, 14\penalty0 (1):\penalty0 6631, March 2024.
\newblock ISSN 2045-2322.
\newblock \doi{10.1038/s41598-024-56674-2}.
\newblock Publisher: Nature Publishing Group.

\bibitem[Mikelsons et~al.(2017)Mikelsons, Smith, Mehrotra, and Musolesi]{mikelsons_towards_2017}
Gatis Mikelsons, Matthew Smith, Abhinav Mehrotra, and Mirco Musolesi.
\newblock Towards {Deep} {Learning} {Models} for {Psychological} {State} {Prediction} using {Smartphone} {Data}: {Challenges} and {Opportunities}.
\newblock In \emph{Proceedings of the {NIPS} {Workshop} on {Machine} {Learning} for {Healthcare} 2017}, November 2017.
\newblock \doi{10.48550/arXiv.1711.06350}.

\bibitem[Miller et~al.(2009)Miller, Vachon, and Lynam]{miller_neuroticism_2009}
Drew~J. Miller, David~D. Vachon, and Donald~R. Lynam.
\newblock Neuroticism, {Negative} {Affect}, and {Negative} {Affect} {Instability}: {Establishing} {Convergent} and {Discriminant} {Validity} {Using} {Ecological} {Momentary} {Assessment}.
\newblock \emph{Personality and individual differences}, 47\penalty0 (8):\penalty0 873--877, December 2009.
\newblock ISSN 0191-8869.
\newblock \doi{10.1016/j.paid.2009.07.007}.

\bibitem[Mishra et~al.(2020)Mishra, Pope, Lord, Lewia, Lowens, Caine, Sen, Halter, and Kotz]{mishra_continuous_2020}
Varun Mishra, Gunnar Pope, Sarah Lord, Stephanie Lewia, Byron Lowens, Kelly Caine, Sougata Sen, Ryan Halter, and David Kotz.
\newblock Continuous {Detection} of {Physiological} {Stress} with {Commodity} {Hardware}.
\newblock \emph{ACM Transactions on Computing for Healthcare}, 1\penalty0 (2):\penalty0 1--30, April 2020.
\newblock ISSN 2691-1957, 2637-8051.
\newblock \doi{10.1145/3361562}.

\bibitem[Mundnich et~al.(2020)Mundnich, Booth, L'Hommedieu, Feng, Girault, L'Hommedieu, Wildman, Skaaden, Nadarajan, Villatte, Falk, Lerman, Ferrara, and Narayanan]{mundnich2020tiles}
Karel Mundnich, Brandon~M. Booth, Michelle L'Hommedieu, Tiantian Feng, Benjamin Girault, Justin L'Hommedieu, Mackenzie Wildman, Sophia Skaaden, Amrutha Nadarajan, Jennifer~L. Villatte, Tiago~H. Falk, Kristina Lerman, Emilio Ferrara, and Shrikanth Narayanan.
\newblock {TILES-2018, a longitudinal physiologic and behavioral data set of hospital workers}.
\newblock \emph{Sci Data}, 7\penalty0 (354), 2020.
\newblock \doi{10.1038/s41597-020-00655-3}.

\bibitem[Neil et~al.(2016)Neil, Pfeiffer, and Liu]{neil_phased_2016}
Daniel Neil, Michael Pfeiffer, and Shih-Chii Liu.
\newblock Phased {LSTM}: {Accelerating} {Recurrent} {Network} {Training} for {Long} or {Event}-based {Sequences}.
\newblock In \emph{Advances in {Neural} {Information} {Processing} {Systems}}, volume~29. Curran Associates, Inc., 2016.

\bibitem[Rubanova et~al.(2019)Rubanova, Chen, and Duvenaud]{rubanova_latent_2019}
Yulia Rubanova, Ricky T.~Q. Chen, and David~K Duvenaud.
\newblock Latent {Ordinary} {Differential} {Equations} for {Irregularly}-{Sampled} {Time} {Series}.
\newblock In \emph{Advances in {Neural} {Information} {Processing} {Systems}}, volume~32, 2019.

\bibitem[Shaw et~al.(2019)Shaw, Simsiri, Deznabi, Fiterau, and Rahaman]{shaw_personalized_2019}
Abhinav Shaw, Natcha Simsiri, Iman Deznabi, Madalina Fiterau, and Tauhidur Rahaman.
\newblock Personalized student stress prediction with deep multi-task network.
\newblock In \emph{Proceedings of the ICML Workshop on Adaptive \& Multitask Learning: Algorithms \& Systems}, May 2019.

\bibitem[Shaw et~al.(2018)Shaw, Uszkoreit, and Vaswani]{shaw_self-attention_2018}
Peter Shaw, Jakob Uszkoreit, and Ashish Vaswani.
\newblock Self-{Attention} with {Relative} {Position} {Representations}.
\newblock In \emph{Proceedings of the 2018 {Conference} of the {North} {American} {Chapter} of the {Association} for {Computational} {Linguistics}}, pages 464--468. Association for Computational Linguistics, June 2018.
\newblock \doi{10.18653/v1/N18-2074}.

\bibitem[Shiffman et~al.(2008)Shiffman, Stone, and Hufford]{shiffman_ecological_2008}
Saul Shiffman, Arthur~A. Stone, and Michael~R. Hufford.
\newblock Ecological {Momentary} {Assessment}.
\newblock \emph{Annual Review of Clinical Psychology}, 4\penalty0 (Volume 4, 2008):\penalty0 1--32, April 2008.
\newblock ISSN 1548-5943, 1548-5951.
\newblock \doi{10.1146/annurev.clinpsy.3.022806.091415}.
\newblock Publisher: Annual Reviews.

\bibitem[Triantafillou et~al.(2019)Triantafillou, Saeb, Lattie, Mohr, and Kording]{triantafillou_relationship_2019}
Sofia Triantafillou, Sohrab Saeb, Emily~G. Lattie, David~C. Mohr, and Konrad~Paul Kording.
\newblock Relationship {Between} {Sleep} {Quality} and {Mood}: {Ecological} {Momentary} {Assessment} {Study}.
\newblock \emph{JMIR Mental Health}, 6\penalty0 (3), March 2019.
\newblock \doi{10.2196/12613}.

\bibitem[Vaswani et~al.(2017)Vaswani, Shazeer, Parmar, Uszkoreit, Jones, Gomez, Kaiser, and Polosukhin]{vaswani_attention_2023}
Ashish Vaswani, Noam Shazeer, Niki Parmar, Jakob Uszkoreit, Llion Jones, Aidan~N Gomez, {\L}ukasz Kaiser, and Illia Polosukhin.
\newblock Attention is {All} you {Need}.
\newblock In \emph{Advances in {Neural} {Information} {Processing} {Systems}}, volume~30, 2017.

\bibitem[Wang et~al.(2014)Wang, Chen, Chen, Li, Harari, Tignor, Zhou, Ben-Zeev, and Campbell]{wang_studentlife_2014}
Rui Wang, Fanglin Chen, Zhenyu Chen, Tianxing Li, Gabriella Harari, Stefanie Tignor, Xia Zhou, Dror Ben-Zeev, and Andrew~T. Campbell.
\newblock {StudentLife}: assessing mental health, academic performance and behavioral trends of college students using smartphones.
\newblock In \emph{Proceedings of the 2014 {ACM} {International} {Joint} {Conference} on {Pervasive} and {Ubiquitous} {Computing}}, {UbiComp} '14, pages 3--14, New York, NY, USA, September 2014. Association for Computing Machinery.
\newblock ISBN 978-1-4503-2968-2.
\newblock \doi{10.1145/2632048.2632054}.

\bibitem[Yu and Sano(2023)]{yu_semi-supervised_2023}
Han Yu and Akane Sano.
\newblock Semi-{Supervised} {Learning} for {Wearable}-based {Momentary} {Stress} {Detection} in the {Wild}.
\newblock \emph{Proceedings of the ACM on Interactive, Mobile, Wearable and Ubiquitous Technologies}, 7\penalty0 (2):\penalty0 1--23, June 2023.
\newblock ISSN 2474-9567.
\newblock \doi{10.1145/3596246}.

\bibitem[Zhang et~al.(2024)Zhang, Jung, Alikhanov, Ahmed, and Lee]{zhang_reproducible_2024}
Panyu Zhang, Gyuwon Jung, Jumabek Alikhanov, Uzair Ahmed, and Uichin Lee.
\newblock A {Reproducible} {Stress} {Prediction} {Pipeline} with {Mobile} {Sensor} {Data}.
\newblock \emph{Proceedings of the ACM on Interactive, Mobile, Wearable and Ubiquitous Technologies}, 8\penalty0 (3):\penalty0 1--35, August 2024.
\newblock ISSN 2474-9567.
\newblock \doi{10.1145/3678578}.

\bibitem[Zhang(2019)]{zhang_attain_2019}
Y.~Zhang.
\newblock {ATTAIN}: {Attention}-based {Time}-{Aware} {LSTM} {Networks} for {Disease} {Progression} {Modeling}.
\newblock \emph{In Proceedings of the 28th International Joint Conference on Artificial Intelligence (IJCAI-2019), pp. 4369-4375, Macao, China.}, January 2019.

\bibitem[Zhu et~al.(2017)Zhu, Li, Liao, Wang, Guan, Liu, and Cai]{zhu_what_2017}
Yu~Zhu, Hao Li, Yikang Liao, Beidou Wang, Ziyu Guan, Haifeng Liu, and Deng Cai.
\newblock What to do next: modeling user behaviors by time-{LSTM}.
\newblock In \emph{Proceedings of the 26th {International} {Joint} {Conference} on {Artificial} {Intelligence}}, {IJCAI}'17, pages 3602--3608, Melbourne, Australia, August 2017. AAAI Press.
\newblock ISBN 978-0-9992411-0-3.

\end{thebibliography}






\end{document}